\documentclass[runningheads]{llncs}

\usepackage{eccv}

\usepackage{eccvabbrv}

\usepackage{graphicx}
\usepackage{booktabs}
\usepackage{multirow}
\usepackage{pifont}
\usepackage[inkscapelatex=false]{svg}
\usepackage{bm}
\usepackage{caption}
\usepackage[accsupp]{axessibility}  

\usepackage{hyperref}

\usepackage{orcidlink}

\begin{document}

\title{Self-Supervised Automatic Matting}


\author{Xiaonan Hu\inst{1} \and
Zhiyuan Lu\inst{2}\and
Jingdong Zhao\inst{3}\and Hao Lu\inst{1, *} 
}
\authorrunning{F.~Author et al.}

\institute{Huazhong University of Science and Technology, China \and
Beijing Normal University, China \and
Waseda University, Japan \\
\email{\{xiaonah, hlu\}@hust.edu.cn}}

\maketitle
\begingroup
\renewcommand{\thefootnote}{\fnsymbol{footnote}}
\footnotetext[1]{Corresponding author.}
\endgroup


\begin{abstract}

High-quality alpha mattes are notoriously expensive to annotate, creating a fundamental data bottleneck for deep image matting. While prior work attempts to reduce annotation cost using coarser labels like trimaps or masks, they remain reliant on costly per-pixel supervision, limiting scalability and generalization. In this work, we push the boundary further and ask: can we train an automatic matting model using only RGB images, with no manual annotation at all? We answer this by presenting SSMatte, a self-supervised framework that for the first time achieves performance on par with fully-supervised automatic matting. Our key insight is to decompose the problem into semantic anchoring and detail matting. SSMatte first generates a semantic matting prompt from frozen self-supervised ViT features by propagating class-token seeds via a novel, training-efficient semantic anchoring loss based on a generalized Rayleigh quotient. This prompt then anchors a detail matting network, which is optimized via a fixed-point-based loss that enforces alpha-RGB consistency. Extensive experiments show SSMatte outperforms prior weakly-supervised methods, matches the performance of fully-supervised models on portrait benchmarks, and demonstrates favorable scaling and generalization behaviors with additional data. Our work pushes automatic matting to an fresh, fully annotation-free paradigm. Code will be available.

  \keywords{Image Matting \and Self-Supervised Learning}
\end{abstract}


\section{Introduction}
\label{sec:intro}
Image matting aims to estimate the per-pixel foreground opacity $\bm \alpha$, a.k.a. alpha matte, from an image $\bm I$, following the matting equation
\begin{equation}
    \bm I=\bm \alpha \bm F + (1-\bm \alpha) \bm B\,,
    \label{equ:matting}
\end{equation}
where $\bm F$ and $\bm B$ are foreground and background, respectively.
It is crucial for applications requiring high-fidelity foreground extraction. While deep learning has revolutionized the field~\cite{xu2017deep}, its success heavily relies on large-scale, high-quality annotations of alpha mattes, which are extremely costly and time-consuming to obtain. The nature of alpha matte, however, makes it almost impossible to annotate/generate at scale, particularly in complex real-world scenarios. This annotation bottleneck severely limits the scalability and generalization of deep matting models.

\begin{figure}[!t]
    \centering
    \includegraphics[width=\textwidth]{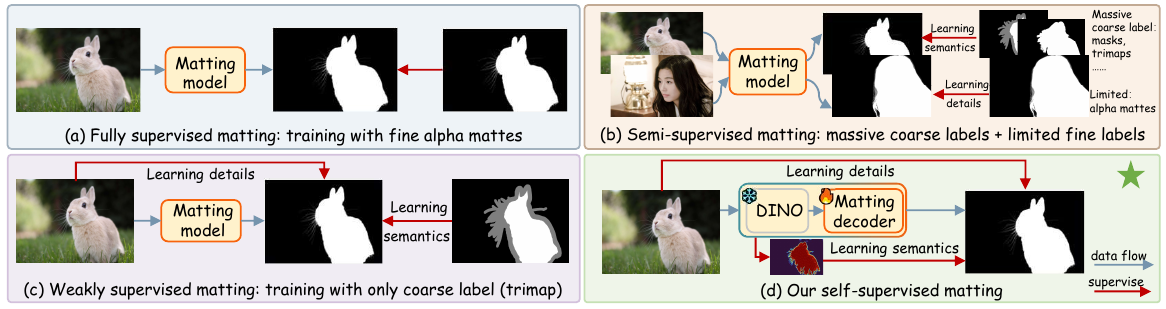}
    \vspace{-15pt}
    \caption{\textbf{Comparison between prior training paradigms of deep matting and our self-supervised matting paradigm}. Prior paradigms require large-scale dense annotations, while we can train an automatic matting model with only RGB images.}
    \label{fig:intro}
\end{figure}

Prior effort to alleviate this bottleneck follows two main strands. The first uses synthetic data~\cite{xu2017deep,qiao2020attention,zhang2019late} or collects subject-specific datasets~\cite{li2022bridging, li2021privacy,sun2023ultrahigh}, yet the domain gap and limited object categories remain issues. The second strand seeks label-efficient supervision, such as training in a semi-supervised manner that uses abundant coarse masks and a few fine alphas~\cite{liu2020boosting,park2023mask,kim2025zim,kong2023semi,zhang2023semi,zhang2023weakly,li2024disentangled,xu2021virtual}. In particular, a notable stride is Alpha-Free Matting (AFM)~\cite{liu2025training}, which eliminates alpha supervision by using only trimaps and a color affinity loss.

However, a critical limitation persists: existing methods still require per-pixel annotations, be it trimaps, masks, or scribbles. Acquiring such annotations, even if coarser than alpha, remains labor-intensive and impedes large-scale data collection. This leads us to a fundamental, unexplored question: \textit{Is it possible to train a competitive automatic matting model with no manual annotation whatsoever, using only RGB images?}

We posit that the answer lies in the synergy between modern self-supervised visual representations and the intrinsic image-matte relationship. Recent self-supervised Vision Transformers (ViTs)~\cite{caron2021emerging,oquab2023dinov2,simeoni2025dinov3} exhibit an emergent capability to localize salient objects without any labels, offering a potential source of ``free'' semantic guidance. Concurrently, the matting equation itself establishes a strong constraint between RGB colors and alpha values. In this work, we bridge these two insights and introduce SSMatte, a self-supervised framework that learns to estimate alpha mattes using only RGB images. 

SSMatte features two key designs: semantic anchoring and detail matting.
For semantic anchoring, SSMatte starts by discovering semantic foregrounds embedded in the self-supervised ViT features~\cite{oquab2023dinov2}. Treating them as seeds, we introduce a semantic anchoring loss $\mathcal{L}_{sea}$ that propagates these seeds across the image by leveraging the intrinsic feature affinities of the pretrained ViT, yielding a coherent, 
soft 
semantic matting prompt 
that effectively replaces a manual trimap. For detail matting, we further derive a target matting loss, a fixed-point-based objective that enforces consistency between the predicted alpha matte and the input RGB image. 
In short, the semantic prompt anchors the global context, telling the model \textit{what} to extract, while the target loss provides a signal for \textit{how} to extract it with high fidelity. By unifying these two self-supervised signals, SSMatte is trained end-to-end, establishing, to our knowledge, the first demonstration of automatic matting without any manual annotations.

Extensive experiments on multiple benchmarks, including P3M~\cite{li2021privacy}, AIM~\cite{li2022bridging}, and AM-2K~\cite{li2021deep} demonstrate that our self-supervised SSMatte not only outperforms 
prior weakly-supervised methods, 
but also achieves competitive performance against state-of-the-art fully-supervised models, particularly on portrait. This closing of the supervision gap is a key result. Moreover, SSMatte shows reasonable generalization: with only lightweight adaptation, it extends 
beyond 
portraits and animals
to diverse object categories. Ablation studies confirm the critical role of both semantic anchoring and detail matting. A further scaling study reveals that performance improves 
with more unlabeled images, highlighting the framework's data efficiency and scalability. Since the ViT backbone remains frozen, training is computationally efficient. Collectively, our work provides the first compelling evidence that \textit{high-quality automatic image matting can be learned from RGB images alone, without any form of manual annotation}.




\section{Related Work}
We review label-efficient deep matting and self-supervised visual learning.

\vspace{-10pt}
\subsubsection{Label-Efficient Deep Matting.}
Since DIM~\cite{xu2017deep} introduced the end-to-end matting paradigm, 
much effort has been made to reduce the dependence on auxiliary inputs and costly annotations.
Early work simplified trimaps into cheaper forms of guidance such as
binary masks~\cite{park2023mask,yu2021mask}, backgrounds~\cite{sengupta2020background,lin2021real}, and user interactions~\cite{wei2021improved}.
A step further, subsequent work aims to 
discard auxiliary guidance and
predict alpha mattes directly from RGB images~\cite{chen2018semantic,li2021privacy,li2021deep,ke2022modnet,li2022bridging}. 
Albeit efficient, 
automatic matting 
suffers from poor generalization, limiting its practical use in specific categories. 
Although some work~\cite{li2021deep,deora2021salient} attempts to 
improve generalization 
by pretraining for salient object detection (SOD)~\cite{wang2017learning}, 
model generalization 
remains limited and relies on 
extra training samples of 
other tasks. 

In response to 
data deficiency, several attempts have driven matting towards 
label-efficient training. 
One main strand 
uses coarse labels (e.g., segmentation masks and trimaps) to reduce the annotation burden~\cite{liu2020boosting,park2023mask,kim2025zim,kong2023semi,zhang2023semi,zhang2023weakly,kim2024towards}. 
These methods distribute abundant semantics in alpha learning for coarse labels, and leave the challenging detailed refinement for fine-grained mattes.
Besides this semi-supervised paradigm, another direction 
devoted to eliminate alpha matte entirely. A successful trial is to train matting models with only trimap supervision~\cite{liu2025training}. 
Albeit reducing dependency on alpha mattes, the methods above still require 
per-pixel 
manual annotations. 
In this work, we challenge a rather provoking setting: train a matting model using only RGB images, without any form of annotations.

\vspace{-10pt}
\subsubsection{Self-Supervised Visual Learning.}

Self-supervised learning (SSL) learns visual representations without human annotations by solving pretext objectives that capture intrinsic structures of images.
Early approaches explored pretext tasks based on image inpainting~\cite{he2022masked,bao2021beit,assran2023self}, while more recent methods introduced contrastive learning, and leveraged discriminative signals between images to learn visual representation~\cite{oquab2023dinov2,caron2021emerging,he2020momentum,simeoni2025dinov3,chen2021exploring,grill2020bootstrap}. For example, DINO~\cite{caron2021emerging} trains a ViT by aligning 
representations of different augmented views of the same image. 
Beyond representation quality, self-supervised ViT features exhibit several emergent properties.
One above all 
is its class-specific property, with features belonging to the same semantic object naturally clustering together without any explicit supervision. 
This property has inspired several unsupervised dense prediction tasks, including localization and segmentation~\cite{simeoni2023unsupervised,melas2022deep,bielski2022move}. However, 
extending SSL from high-level dense prediction to fine-grained matting
remains largely unexplored. 
Existing attempts related to self-supervised matting are 
designed primarily for localization/segmentation~\cite{melas2022deep} and provide limited validation on standard matting benchmarks.
Consequently, a challenging gap still 
exists in bridging self-supervised features with per-pixel image matting. 

\begin{figure}[!t]
    \centering
    \includegraphics[width=\textwidth]{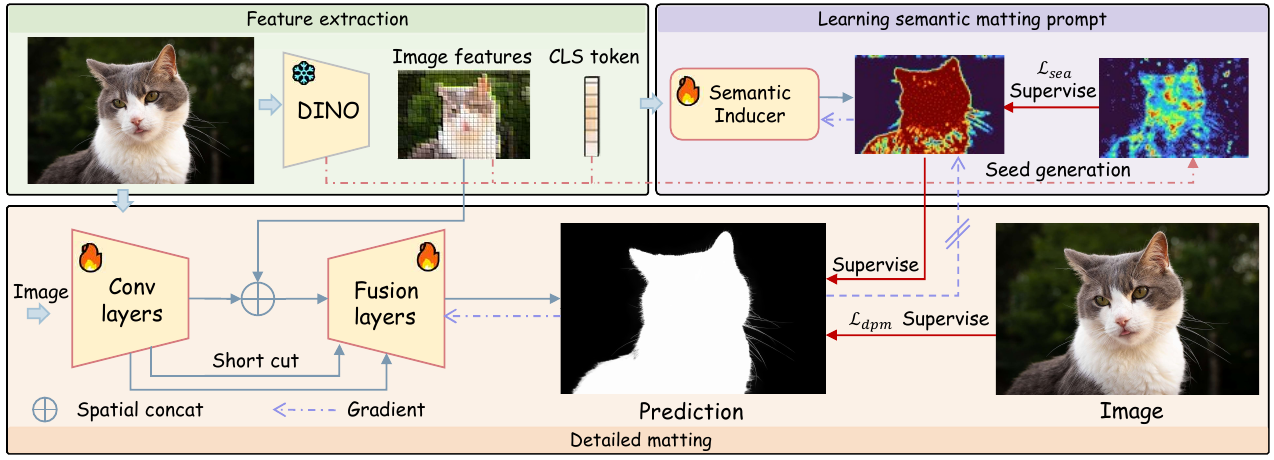}
    \vspace{-15pt}
    \caption{\textbf{Technical pipeline of our self-supervised automatic image matting}. We mine a prompt from frozen DINO to anchor semantics for subsequent matting, and utilize the alpha-RGB consistency to refine details.}
    \label{fig:pipeline}
\end{figure}

\section{Self-Supervised Automatic Matting}


In this section, we present an end-to-end framework for self-supervised automatic matting. 
Per Fig.~\ref{fig:pipeline}, our key insight is to 
seek a high-level matting prompt, anchoring semantics for 
low-level detail matting. We begin by analyzing the foreground cues emerged in self-supervised ViT features. To provide a stable and coherent semantic anchor for matting, 
we 
leverage patch affinities to
propagate 
the matting prompt 
in a self-supervised manner (Sec.~\ref{sec:sfg}). With this 
prompt and texture cues in RGB images, we 
then introduce how to infer alpha mattes 
via RGB-alpha consistency (Sec.~\ref{sec:dm}).

\subsection{Self-Supervised 
Foreground Prompting}\label{sec:sfg}
\subsubsection{Foreground Seed.}\label{sec:seed}
Without annotations, the fundamental challenge in self-supervised image matting is to determine which object to extract. In this work, we follow the setting of automatic image matting and further prioritize salient 
opaque 
foregrounds.
In automatic matting, much effort has been made to recognize the matting object~\cite{li2021deep,deora2021salient}. A recent 
example is Smat~\cite{ye2024unifying}, where 
candidate object features 
are constructed based on the \texttt{[CLS]} token in ViT. The \texttt{[CLS]} token is originally designed for image-level pretexts like class assignment, which relies on the most salient object in images. In self-supervised ViTs, the foreground cues naturally emerge on the attention map between the \texttt{[CLS]} token and patch features~\cite{caron2021emerging}, manifested as high responses 
in a cosine similarity map. 

\begin{figure}[!t]
    \centering
    \includegraphics[width=\linewidth]{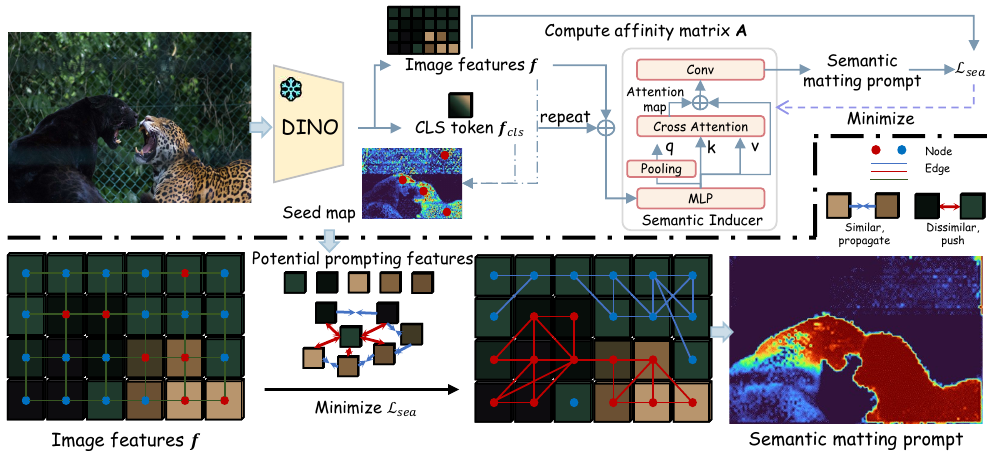}
    \vspace{-15pt}
    \caption{\textbf{Details of the proposed Semantic Inducer and $\mathcal{L}_{sea}$}. $\mathcal{L}_{sea}$ encourages similar patch nodes to have similar response values in the predicted semantic matting prompt. We supervise a cross attention layer to obtain the semantic matting prompt.}
    \label{fig:lsea}
\end{figure}

We extend this property into a fully label-free context for image matting.
Formally, given an image $\bm I\in \mathbb{R}^{3\times H\times W}$ of resolution $H \times W$, a ViT $\Phi_{enc}$ extracts image features $\bm f\in \mathbb{R}^{d\times \frac{H}{p} \times \frac{W}{p}}$ and a \texttt{[CLS]} token $\bm f_{cls} \in \mathbb{R}^{d\times 1}$, where $d$ is the feature dimension, and $p$ is the patch size. We define a foreground seed map by
\begin{equation}
    \bm S_0=\mathrm{MinMaxNorm}(\frac{1}{E}\sum_{e=1}^E\mathrm{sim}(\bm q^{i,e}_{cls}, \bm k^{i,e}_{feat})),\,
    \label{equ:seed}
\end{equation}
where $i$ denotes the $i$-th layer of $\Phi_{enc}$, $e$ denotes the $e$-th of the total $E$ attention heads, $\bm q_{cls}$ and $\bm k_{feat}$ denote the query and key vector of \texttt{[CLS]} token and image features, $\mathrm{MinMaxNorm}$ denotes the min-max normalization, and $\mathrm{sim}$ denotes the cosine similarity, respectively. 
As shown in Fig.~\ref{fig:lsea}, this seed map roughly outlines the foreground with high response values, namely \textit{seed} patches. 

\vspace{-10pt}
\subsubsection{Merge Patch Affinities for Matting Prompting.}
The sparsity and noise in the foreground seeds indicate that the \texttt{[CLS]} token alone is insufficient for a matting prompt. To mine a semantically coherent matting prompt, we need supervision to connect isolated seeds of high response values. 
A recent study revealed that self-supervised ViTs emerge object binding ability: they can capture whether two patches belong to the same object via similarity between image patches~\cite{li2025does}. 
Inspired by this emergent property, we seek to \textit{propagate} the sparse and noisy seed based on the affinities between self-supervised ViT features. 
We first construct a similarity matrix in the feature space, formulated by
\begin{equation}
    \bm{A}=\mathrm{max}(\bm{f}^T\bm{f},0) \in \mathbb{R}^{N\times N},\,
\end{equation}
where $N=\frac{H}{p}\times \frac{W}{p}$ denotes the patch numbers, $\bm f$ is flattened to $d\times N$, and $\mathrm{max}$ is applied element wise. With these affinities off-the-shelf, we assume that a semantic prompt for matting should contain patches that are highly similar to seed patches, as many as possible.  
We formulate this assumption by minimizing
\begin{equation}
    \mathcal{L}=\frac{\hat{\bm S}^T \bm L\hat{\bm S}}{\hat{\bm S}^T\bm D\hat{\bm S}}+\frac{(\bm 1_N-\hat{\bm S})^T\bm L(\bm 1_N-\hat{\bm S})}{(\bm 1_N-\hat{\bm S})^T\bm D(\bm 1_N-\hat{\bm S})} + \lambda_1||\hat{\bm S}-\hat{\bm S}_0||_2,\,
    \label{equ:cut}
\end{equation}
where $\hat{\bm S}= \{0,1\}^{N \times 1}$ is the binarized and flattened matting prompt, $\bm 1_N$ denotes the $N$-dimensional homogeneous column, $\bm D$ denotes the diagonal degree matrix entering at $\bm D_{ii}=\sum_{j=1}^N \bm A_{ij}$, and $\bm L = \bm D - \bm W$ denotes the Laplacian matrix. 
Subsequently, we introduce a $\bm D-$weighted mean $\mu=\frac{\hat{\bm S}^T \bm{D} \bm{1}_N}{\bm{1}_N^T \bm{D} \bm{1}_N}$, and perform a centering operation on $\hat{\bm S}$ by $\bm{u}=\hat{\bm S}-\mu \bm{1}_N$. Thus, we can rewrite Eq.~\eqref{equ:cut} as
\begin{equation}
    \mathcal{L}=\frac{\bm u^T\bm L\bm u}{\bm u^T\bm D\bm u}+\lambda_1 ||\bm u-\bm u_0||_2,\,
    \label{equ:ucut}
\end{equation}
Notice that the seed loss term $||\bm u-\bm u_0||_2$ is essentially an unweighted Euclidean distance across the whole seed map, while we only seek to take in patches that are highly similar to the seed patches. Considering $\bm D$ is diagonal positive definite, 
we thus replace $||\bm u-\bm u_0||_2$ with its weighted counterpart $(\bm u-\bm u_0)^T\bm D(\bm u-\bm u_0)$, which driving the centralized prompt $\bm u$ towards the centralized seed $\bm u_0$. 
Guided by this insight, we reformulate Eq.~\eqref{equ:ucut} as
\begin{equation}
    \mathcal{L}=\frac{\bm u^T\bm L\bm u}{\bm u^T\bm D\bm u}+\lambda_1(\bm u-\bm u_0)^T\bm D(\bm u-\bm u_0)=\frac{\bm u^T\bm L\bm u}{\bm u^T\bm D\bm u}+\lambda_1 (\bm u^T\bm D\bm u-2\bm u_0^T\bm D\bm u)+\mathrm{const}.
    \label{equ:ucutseed}
\end{equation}
Notice that $\frac{\bm u^T\bm L\bm u}{\bm u^T\bm D\bm u}$ is a generalized Rayleigh quotient, which is scale‑invariant with respect to $\bm u$. Motivated by this property, we decompose $\bm u$ into a direction vector $\bm v$ and a scale factor $\rho$ as $\bm u=\rho \bm v$, and rewrite Eq.~\eqref{equ:ucutseed} as
\begin{equation}
    \mathcal{L}(\bm u) = \mathcal{L}(\rho,\bm v)= \frac{\bm v^T\bm L\bm v}{\bm v^T\bm D\bm v} + \lambda_1 (\bm v^T\bm D\bm v \rho^2 - 2\bm u_0^T \bm D \bm v \rho)
    \label{equ:lrouv}
\end{equation}
Eq.~\eqref{equ:lrouv} gives the optimal scale $\rho=\frac{\bm u_0^T\bm D\bm v}{\bm v^T\bm D\bm v}$ and thus can be simplified into a generalized Rayleigh quotient as 
\begin{equation}
    \mathcal{L}=\frac{\bm v^T(\bm L-\lambda_1 \bm D\bm u_0\bm u_0^T\bm D)\bm v}{\bm v^T\bm D\bm v}.
    \label{equ:rayligh}
\end{equation}
Empirically, the second smallest eigenvector of the corresponding eigenvalue system $(\bm L-\lambda \bm D\bm u_0^T\bm u_0\bm D)\bm v=\mu \bm D\bm v$ can be the centralized semantic matting prompt~\cite{shi2000normalized,wang2023tokencut}, where $\mu$ denotes the eigenvalue. For more discussions, please refer to the Supplementary. 

\vspace{-10pt}
\subsubsection{Semantic Inducer.}
Rather than solving a linear system for each input image, we employ a network to generate the soft matting prompt $\bm S\in[0,1]^{N\times 1}$. 
Specifically, we flatten the self-supervised features $\bm f\in \mathbb{R}^{d\times \frac{H}{p} \times \frac{W}{p}}$ into a $d\times N$ matrix, and construct a query vector by a global pooling~\cite{guan2025contrastive} as $\bm q = \mathrm{softmax}(\bm f)^T\bm f$,
where the $\mathrm{softmax}$ is operated along the spatial channel. We then apply a cross-attention layer with $\bm k=\mathrm{MLP}(\mathrm{concat}(\bm f, \mathrm{repeat}(\bm f_{cls}))$ as the key and value vector, where $\mathrm{repeat}$ denotes repeat the \texttt{[CLS]} token for $N$ times to apply a dimensional concat, and get the attention map $\bm{attn}$ to obtain the final prediction, formulated by $\bm S=\mathrm{Conv2D}(\mathrm{concat}(\bm{attn},\bm f))$.
To avoid the prediction from collapsing into ambiguity, we further add a regularization term $\mathcal{L}_{orth}=||\frac{\bm \phi(\bm u)^T \phi(\bm u)}{||\phi(\bm u)^T \phi(\bm u)||_F } -\frac{\bm I}{\sqrt{2}}||_F$~\cite{bianchi2020spectral}, where $\phi(\bm u)=[\bm u, \bm 1_N-\bm u]$, and relax $\bm D\bm u_0 \bm u_0^T \bm D$ to its diagonal $Tr(\bm D\bm u_0 \bm u_0^T \bm D)$. The final loss function can thus be formulated by
\begin{equation}
    \mathcal{L}_{sea}=\frac{\bm v^T(\bm L-\lambda _1Tr(\bm D\bm u_0\bm u_0^T\bm D))\bm v}{\bm v^T\bm D\bm v} + \lambda_2 ||\frac{\bm \phi(\bm u)^T \bm \phi(\bm u)}{||\bm \phi(\bm u)^T \bm \phi(\bm u)||_F } -\frac{\bm I}{\sqrt{2}}||_F.
    \label{equ:sfm}
\end{equation}
Notably, we freeze $\Phi_{enc}$, and optimize only the parameters of the Semantic Inducer. 
As illustrated in Fig.~\ref{fig:lsea}, this loss function anchors features of the matting target and pushes irrelevant features away, encouraging a soft semantic prompt for matting. For more details, please refer to the Supplementary.  

\subsection{Detail Matting}\label{sec:dm}
\begin{figure}[!t]
    \centering
    \includegraphics[width=\linewidth]{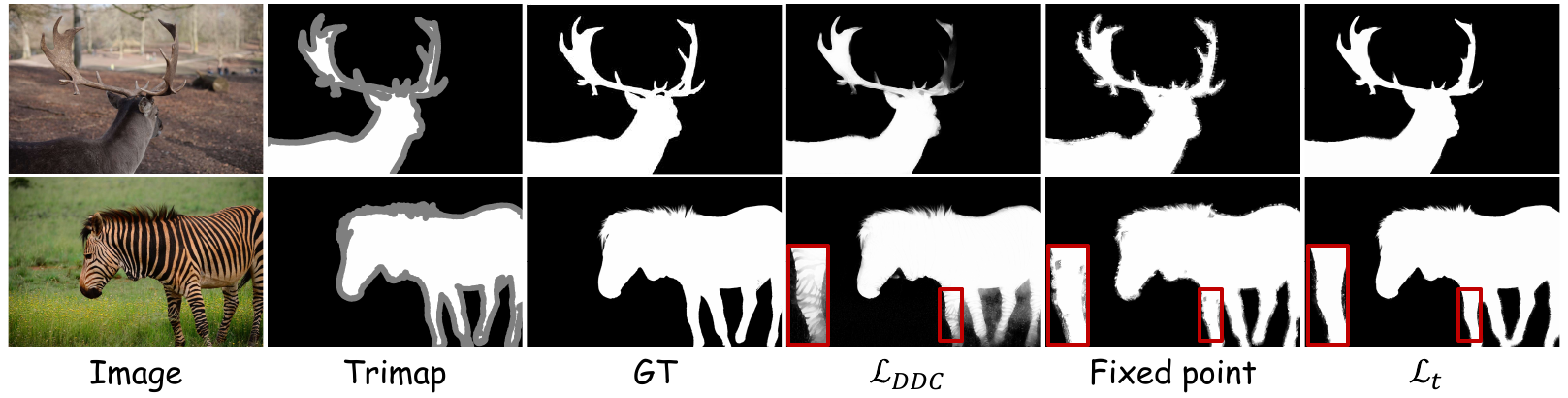}\vspace{-5pt}
    \caption{\textbf{$\mathcal{L}_{DDC}$ accumulates textures during training.} $\mathcal{L}_{DDC}$ denotes the output of alpha-free matting, the estimated fixed point denotes the converged solution, and $\mathcal{L}_{t}$ denotes supervising alpha-free matting with $\mathcal{L}_{t}$.}
    \label{fig:ddc}
\end{figure}

\subsubsection{Preliminary: Directional Distance Consistency.}
With the matting prompt serving as semantic anchors, we still need to find supervising signals for predicting detailed alpha mattes. 
A recent work~\cite{liu2025training} showcased how to 
learn detailed alpha from RGB images. 
Inspired by the nonlocal principle in traditional matting, the proposed method, alpha-free matting(AFM), assumed that pixels sharing similar colors should share consistently similar alpha mattes, formulated by the Directional Distance Consistency(DDC) loss as 
\begin{equation}
    \mathcal{L}_{DDC}=\frac{1}{M}\sum_i^M\sum_j^K|\bm \alpha_i-\bm \alpha_j-||\bm I_i-\bm I_j||_2|,j\in \text{argtopk}\{-||\bm I_i-\bm I_j||_2\},\,
\end{equation}
where $M$ denotes the pixel numbers, and $K$ denotes the selected similar pixel numbers. In their work, they leverage trimaps for learning semantics, and utilize $\mathcal{L}_{DDC}$ to propagate alpha learned in foreground/background regions iteratively.

\vspace{-10pt}
\subsubsection{Merge Semantic Prompt for Detail Matting.}\label{sec:lt}
The original DDC loss 
utilizes trimaps as semantic anchors for known and unknown regions, 
an input that is unavailable in our self-supervised setting. 
A naive solution is to replace the trimap with a pseudo one,
which can be morphologically eroded from the semantic matting prompt. 
Unfortunately, this replacement is proven to be problematic. 
Even with a GT trimap, $\mathcal{L}_{DDC}$ can still introduce certain unexpected alpha textures near boundaries, as the boundary region inevitably mixes truly transparent pixels with foreground/backgrounds. 
These undesired variations attribute to the propagation of image textures. 
Mathematically, $\mathcal{L}_{DDC}$ stimulates an ideal propagation as
\begin{equation}
    \bm \alpha{(t+1)}=\bm W \bm \alpha (t) + \bm b,\,
    \label{equ:propagete}
\end{equation}
where $\bm \alpha \in \mathbb{R}^{M\times 1}$ denotes the flattened alpha matte, $t$ denotes the optimizing iteration, $\bm b \in \mathbb{R}^{M\times 1}$ denotes the sum of image textures, entering at $\bm b_i=\frac{1}{K}\sum_j^K||\bm I_i-\bm I_j||_2$, and $\bm W\in \mathbb{R}^{M\times M}$ denotes a correlation matrix, which takes the form of
\begin{equation}
    \bm W_{ij}=\begin{cases}
        \frac{1}{K}, &\text{if } j\in \text{argtopk}\{-||I_i-I_j||_2\}\\
        0,& \text{else}.
    \end{cases}
    \label{equ:linked matrix}
\end{equation}
Eq.~\eqref{equ:propagete} assumes $\bm \alpha(t)$ as an intermediate variable of a fixed iterative solver. However, in the deep learning paradigm, $\bm \alpha(t)$ is actually the output of the model 
parameterized by $\theta_t$, complicating the propagation into 
$\bm \alpha(\theta_{t+1})=\bm W\bm \alpha(\theta_t) + \bm b$. 
Consequently, minimizing $\mathcal{L}_{DDC}$ introduces a path dependency effect. Once the image textures are learned within opacity of foreground/backgrounds in boundaries, $\bm \alpha_i(\theta_t)$, they tend to be preserved and further propagated across subsequent iterations, which accumulates artefacts in foreground regions, as shown in Fig.~\ref{fig:ddc}. 
We further simulate the propagation in Eq.~\eqref{equ:propagete} to isolate this effect. 
Specifically, we iterate Eq.~\eqref{equ:propagete} to convergence, with $\bm \alpha(0)$ initialized as the ground truth trimap. During iteration, we explicitly anchor known regions with the trimap, and clamp the intermediate $\bm \alpha(t)$ to $[0,1]$.  
As shown in Fig.~\ref{fig:ddc}, Col. 5, the converged solution shows significantly fewer textures than the result produced by $\mathcal{L}_{DDC}$. Albeit noisy near ambiguous boundaries, 
this fixed-point output suggests that the foreground textures mainly arise from path-dependent optimization. 
This problem becomes more severe in our setting, as boundary regions in our matting prompt are less reliable than those of a GT trimap.

This observation motivates us to bypass intermediate propagation states. 
Iterating Eq.~\eqref{equ:propagete} to convergence actually amounts to solving the fixed-point condition
$\bm \alpha=\bm W \bm \alpha + \bm b$. 
We therefore introduce a target loss $\mathcal{L}_t=\|\bm{\alpha}-\bm{W}\bm{\alpha}-\bm{b}\|_1$ to encourage a direct prediction of the fixed point. 
To validate its effectiveness, we directly replace $\mathcal{L}_{DDC}$ with $\mathcal{L}_{t}$ in AFM, and randomly relax $10\%$ foregrounds to unknown regions in the trimap during training to simulate a pseudo trimap. 
As shown in Fig.~\ref{fig:ddc}, compared with $\mathcal{L}_{DDC}$, 
$\mathcal{L}_t$ encourages a significantly smoother alpha prediction, especially in boundaries. 
However, it also encounters a slight hard segmentation effect, as already discussed in~\cite{liu2025training}.    
Thus, we formulate the overall detail matting loss by
\begin{equation}
    \mathcal{L}_{dpm}=\mathcal{L}_{sem} + \lambda_3 \mathcal{L}_{t} + \lambda_4 \mathcal{L}_{DDC},\,
    \label{equ:dpm}
\end{equation}
where $\mathcal{L}_{sem}$ is a standard $\mathcal{L}_1$ loss applied only on the high-confidence regions derived from the tri-valued semantic matting prompt $\tilde{\bm S}$, which is formulated by
\begin{equation}
    \tilde{\bm S}_i=\begin{cases}
        1,\text{if}~\mathrm{erode}(\bm S)_i>\theta_1\\
        0,\text{if}~\mathrm{erode}(\bm S)_i<\theta_2\\
        0.5,\text{else},\,
    \end{cases}
    \label{equ:psd trimap}
\end{equation}
where $\mathrm{erode}$ denotes the morphological erosion, and $\bm S$ is directly upsampled. 
In this final objective, $\mathcal{L}_t$ complements $\mathcal{L}_{DDC}$ by directly regularizing the converged solution, while retaining local directional consistency.
For more details about our detail matting loss, please refer to our Supplementary.

\vspace{-10pt}
\subsubsection{Network Optimization.}
Bridging $\mathcal{L}_{sea}$ and $\mathcal{L}_{dpm}$ together we introduce the overall pipeline of our method. Given an image $\bm I$ along with a self-supervised ViT $\Phi_{enc}$, we first apply a Semantic Inducer supervised by $\mathcal{L}_{sea}$ to obtain the semantic matting prompt. Subsequently, we leverage the matting prompt for the detail matting, generating the tri-valued prompt to supervise a matting model, and obtain the final alpha mattes. We formulate the total loss as 
\begin{equation}
    \mathcal{L} =\mathcal{L}_{sea} + \mathcal{L}_{dpm}.
    \label{equ:total}
\end{equation}
We utilize this loss function for an end-to-end self-supervised training.

\section{Experiments}
\subsection{Implementation Details}
\subsubsection{Dataset and Metrics.}
We select the natural image matting datasets AM-2K~\cite{li2022bridging} and P3M-10K~\cite{li2021privacy} for model training, as the composition images interrupt the object capturing ability of the \texttt{[CLS]} token~\cite{ye2024unifying}. We also include a more challenging benchmark AIM-500~\cite{li2021deep}, to test the generalization ability to categories and foreground types of our method. 
We report Sum of Absolute Difference (SAD), Mean Absolute Difference (MAD)~\cite{li2022bridging}, Mean Squared Error (MSE), Gradient Loss (Grad),
and Connectivity Loss (Conn)~\cite{rhemann2009perceptually}. SAD is scaled by $10^3$, MAD, MSE are scaled by $10^2$, and Grad and Conn are scaled by $10^{-3}$. 

\vspace{-10pt}
\subsubsection{Training Details.}
We employ ViTMatte~\cite{yao2024vitmatte} as the deep matting model. Specifically, we choose DINOv2~\cite{oquab2023dinov2} as the pretrained backbone, and freeze its parameters during training. 
We empirically set the selected similar pixels $K$ in Eq.~\eqref{equ:linked matrix} to $11$, $\lambda_1,\lambda_2$ in Eq.~\eqref{equ:sfm} to $0.1$ and $1.5$, $\lambda_3,\lambda_4$ to $20$ and $10$ in Eq.~\eqref{equ:dpm}, and thresholds $\theta_1,\theta_2$ in Eq.~\eqref{equ:psd trimap} to $0.8$ and $0.2$, respectively. We take a batch size of $8$, and train the model for $30$K iterations. 
The entire training procedure takes around $4$ hours on one single NVIDIA A6000. We employ AdamW as the optimizer, with learning rate initialized as $5e^{-4}$ and a decay rate of $0.01$. For more details and analysis on parameter sensitivity, please refer to the Supplementary.

\begin{table}[!t]
\caption{\textbf{Automatic matting results}. $^*$ denotes that the backbone is initialized with DINOv2, and $^\dagger$ denotes that a trimap-guided crop is applied to ensure the image patches contain unknown regions. Best performances and second performances of non-full supervision are in \textbf{boldface} and \underline{underline}, respectively.}
\vspace{-10pt}
\centering
\small
\setlength{\tabcolsep}{1pt}        
\renewcommand{\arraystretch}{1.1}   
\resizebox{\textwidth}{!}{%
\begin{tabular}{lccccccccccccc}
\toprule
\multirow{2}{*}{Method} &
\multirow{2}{*}{Label} &
\multicolumn{6}{c}{AM-2K} &
\multicolumn{6}{c}{P3M-NP-500} \\
\cmidrule(lr){3-8}\cmidrule(lr){9-14}
& & SAD & MAD & MSE & Grad & Conn & SAD-T 
& SAD & MAD & MSE & Grad & Conn & SAD-T \\
\midrule
SHM~\cite{chen2018semantic}  & Matte & 17.81 & 1.02 & 0.68 & 12.54 & 17.02 & 10.26 & 20.77 & 1.22 & 9.3 & 20.30 & 17.09 & 9.14 \\
LF~\cite{zhang2019late}   & Matte & 36.12 & 2.10 & 1.16 & 21.06 & 33.62 & 19.68 & 32.59 & 1.88 & 1.31 & 31.93 & 19.50 & 14.53\\
HATT~\cite{qiao2020attention} & Matte & 28.01 & 1.61 & 0.55 & 18.29 & 17.76 & 13.36 & 30.53 & 1.76 & 0.72 & 19.88 & 27.42 & 13.48\\
SSS~\cite{aksoy2018semantic}  & Matte & 552.88 & 32.25 & 27.42 & 60.51 & 555.97 & 88.23 & \ & \ & \ & \ & \ & \ \\
MODNet~\cite{ke2022modnet} & Matte & \ & \ & \ & \ & \ & \ & 16.70 & 0.97 & 0.51 & 15.29 & 13.91 & 9.13\\
GFM~\cite{li2022bridging} & Matte & 10.26 & 0.59 & 0.29 & 8.82 & 9.57 & 8.24 & 15.50 & 0.91 & 0.56 & 14.82 & 18.03 & 10.16\\
P3M~\cite{li2021privacy}   & Matte & \ & \ & \ & \ & \ & \ & 11.23 & 0.65 & 0.35 & 10.35 & 12.51 & 5.32\\
Smat~\cite{ye2024unifying}  & Matte & 16.84 & 0.98  & 0.46 & 17.97 & 14.16 & 14.90 & 18.93 & 1.10 & 0.64 & 19.07 & 16.35 & 14.09\\
ViTMatte-S~\cite{yao2024vitmatte}  & Matte & 15.53 & 0.86 & 0.58 & 8.12 & 6.20 & 7.86 & 19.56 & 1.13 & 0.84 & 12.22 & 9.82 & 7.30\\
\midrule
AFM~\cite{liu2025training}  & Trimap & 30.26 & 1.75 & 0.99 & \textbf{14.98} & 14.33 & 16.07 & 31.66 & 1.82 & 1.26 & \underline{15.41} & 13.66 & 12.03 \\
AFM$^*$~\cite{liu2025training}  & Trimap & 27.37 & 1.60 & 0.84 & 17.36 & 15.26 & 19.83 & 26.28 & 1.53 & 1.07 & 21.16 & 13.64 & 14.71 \\
SSMatte (Ours)         & RGB    & \underline{25.84} & \underline{1.50} & \underline{0.58} & 17.80 & \underline{11.17} & \underline{16.92} & \underline{17.17} & \underline{0.99} & \underline{0.45} & 18.73 & \underline{8.99} & \underline{11.42}\\
SSMatte$^\dagger$ (Ours)     & RGB    & \textbf{23.54} & \textbf{1.36} & \textbf{0.58} & \underline{15.10} & \textbf{10.35} & \textbf{15.49} & \textbf{13.64} & \textbf{0.79} & \textbf{0.34} & \textbf{15.18} & \textbf{8.11} & \textbf{10.46}\\
\bottomrule
\end{tabular}%
}
\label{tab:main result}
\end{table}

\begin{figure}[!t]
    \centering
    \includegraphics[width=\textwidth]{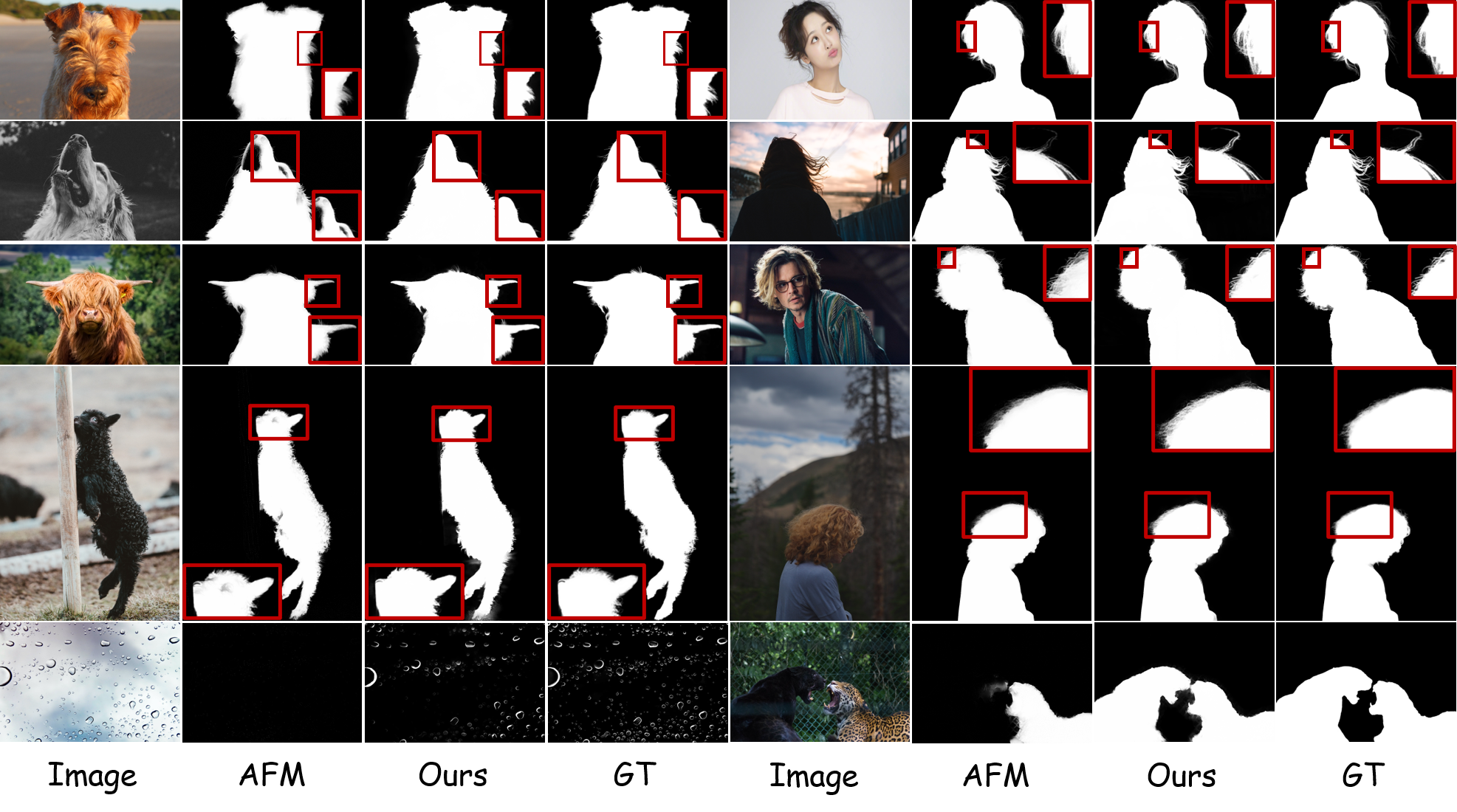}
    \caption{\textbf{Qualitative automatic results on AM-2K and P3M-NP-500}. Our method produces better alpha mattes than AFM at details and recognizing foregrounds, and occasionally produces alpha mattes that are more plausible than ground truths.}
    \label{fig:main result}
\end{figure}

\subsection{Main Results}

We compare our method with fully-supervised automatic methods and a weakly supervised method.  
We also include ViTMatte~\cite{yao2024vitmatte}, excluding the auxiliary channel for trimap, and retraining it purely on AM-2K and P3M-10K, respectively. 
Among baselines, alpha-free matting (AFM)~\cite{liu2025training} is the only method that requires no alpha mattes, making it the most relevant baseline for our method. We also evaluate AFM initialized with DINOv2~\cite{oquab2023dinov2} for a fair comparison. All methods are evaluated in the automatic matting setting.

As shown in Table~\ref{tab:main result}, 
SSMatte outperforms the weakly supervised AFM across most metrics on both benchmarks, experiencing only a marginal decrease in the Grad on AM-2K. 
This minor reduction may stem from the choice of backbones, as 
AFM initialized with DINOv2 improves SAD, MAD, and MSE, while also degrading Grad and Conn compared with its DINOv1 counterpart. 
Compared with fully supervised methods, SSMatte yields performance comparable to MODNet~\cite{ke2022modnet} on the portrait dataset and surpasses HATT~\cite{qiao2020attention} on the animal dataset. 
To test the upper bound of our method, we employ a trimap-guided image cropping strategy to ensure that the training patches contain boundary regions, which is used in several baselines~\cite{yao2024vitmatte,liu2025training,li2022bridging,li2021privacy}. 
Note that trimaps in this strategy are only used 
for data augmentation, rather than for supervising. 
As shown in the last row in Table~\ref{tab:main result}, equipped with this data augmentation, our method outperforms GFM~\cite{li2022bridging} and achieves performance comparable with P3M~\cite{li2021privacy} on the portrait benchmark. 

\begin{table*}[t!]
\centering
\caption{\textbf{Loss ablations}. Each term shows a performance increase across both datasets. Best performances are in \textbf{boldface}.}
\vspace{-10pt}
\small
\resizebox{\textwidth}{!}{
\begin{tabular}{cccccccccccccccc}
\toprule
\multirow{2}{*}{$\mathcal{L}_{sea}$} & \multirow{2}{*}{$\mathcal{L}_{seed}$} & \multirow{2}{*}{$\mathcal{L}_{DDC}$} & \multirow{2}{*}{$\mathcal{L}_{t}$} & \multicolumn{6}{c}{AM-2K} & \multicolumn{6}{c}{P3M-10K}\\
 \cmidrule(lr){5-10}\cmidrule(lr){11-16}
 & & & &  SAD & MAD & MSE & Grad & Conn & SAD-T & SAD & MAD & MSE & Grad & Conn & SAD-T\\
\midrule
\ding{51} &  & \ding{51} & \ding{51} & \textbf{23.54} & \textbf{1.36} & \textbf{0.58} & \textbf{15.10} & \textbf{10.35} & \textbf{15.49} & \textbf{13.64} & \textbf{0.79} & \textbf{0.34} & \textbf{15.18} & \textbf{8.11} & \textbf{10.46}
\\
          & \ding{51} & \ding{51} & \ding{51} & 31.38 & 1.82 & 0.68 & 17.95  & 11.60  & 18.38 & 503.82 & 29.12 & 16.2 & 365.96 & 44.27  & 61.92\\
\ding{51} &  & \ding{51} &           & 46.64 & 2.70 & 0.96 & 17.48 & 33.63 & 28.17 & 103.25 & 6.03 & 4.5 & 20.03 & 55.80 & 21.33\\
\ding{51} &  &           & \ding{51} & 186.42 & 10.83 & 10.19 & 89.87 & 40.86 & 89.95 & 177.14 & 10.29 & 9.7 & 94.24 & 50.26 & 79.70\\
\bottomrule
\end{tabular}%
}
\label{tab:ablation loss}
\end{table*}

\begin{figure}[!t]
    \centering
    \vspace{-10pt}
    \includegraphics[width=\textwidth]{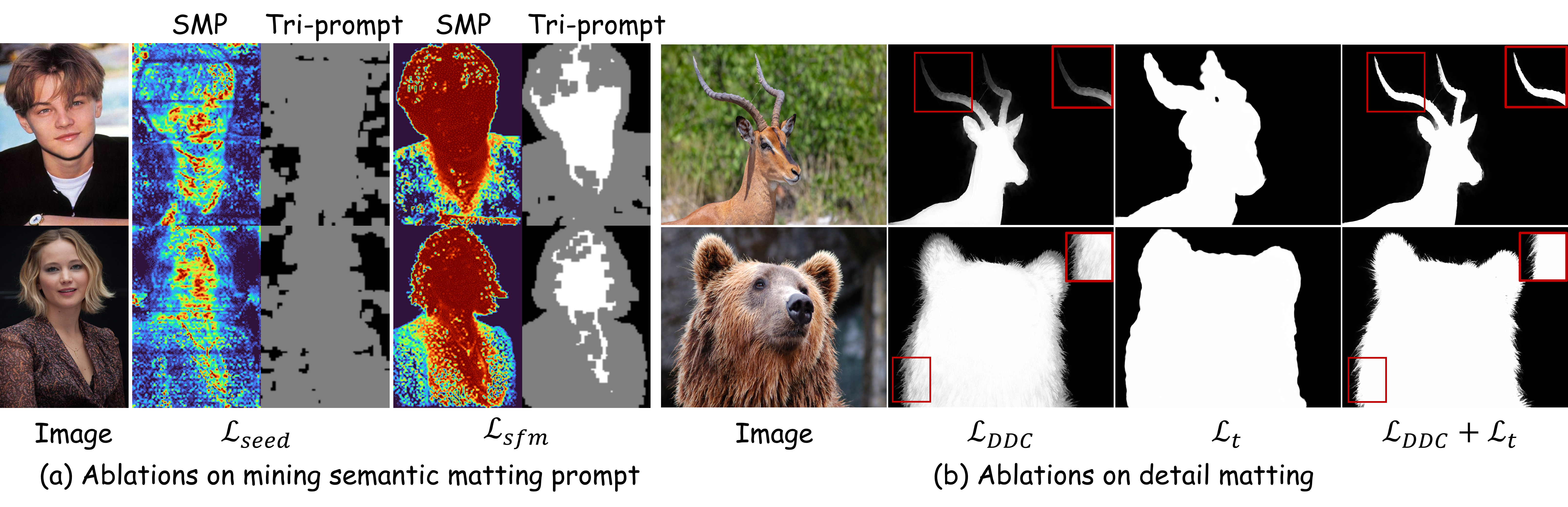}
    \caption{\textbf{Qualitative results of ablations on (a) Mining semantic matting prompt, and (b) Detail matting}. 
    SMP is short for semantic matting prompt. $\mathcal{L}_{seed}$ produces a noisy foreground cue, while $\mathcal{L}_{sea}$ provides a more sound semantic matting prompt. 
    $\mathcal{L}_{DDC}$ forces image textures in boundaries, while $\mathcal{L}_{t}$ encourages a hard segmentation. Only by combining $\mathcal{L}_{DDC}$ and $\mathcal{L}_{t}$ together produces optimal alpha mattes.}
    \label{fig:ab}
\end{figure}

Qualitative results can be seen in Fig.~\ref{fig:main result}. Notably, since our method explicitly incorporates alpha-RGB priors, it occasionally produces alpha mattes that are more visually plausible than ground truths, as shown in the Row. 1, Col. 2, and Row. 4, Col. 2 in Fig.~\ref{fig:main result}. Additionally, benefiting from self-supervision, our method recognizes the target object better than explicitly supervising semantics, as shown in the Row. 1, Col. 1, and Row. 2, Col. 1, and Row. 4, Col. 4 in Fig.~\ref{fig:main result}.


\subsection{Ablation Study}\label{sec:ablation}

Here we evaluate our proposed losses. We train three model variants on P3M-10K and AM-2K: (1) replace $\mathcal{L}_{sea}$ with $\mathcal{L}_{seed}$, (2) remove $\mathcal{L}_{DDC}$, and (3) remove $\mathcal{L}_{t}$. $\mathcal{L}_{seed}$ here denotes $\mathcal{L}_{seed}=||\bm S-\bm S_0||_2$. Note that all variants are trained with the trimap-guided cropping strategy for a fair comparison. As shown in Table~\ref{tab:ablation loss}, each loss term yields a consistent improvement across all metrics. 
For prompt mining, 
models supervised by $\mathcal{L}_{seed}$ perform unevenly across different datasets, yielding reasonable results on AM-2K but poor results on P3M-10K. This inconsistent performance indicates that the naive $\mathcal{L}_{seed}$, which is derived from the \texttt{[CLS]} token, is insufficient for a robust semantic anchoring. 
Quantitative results of ablations on $\mathcal{L}_{sea}$ and $\mathcal{L}_{seed}$ can be seen in Fig.~\ref{fig:ab} (a), where the \texttt{[CLS]} token alone encourages an informative yet noisy and averaged prompt.  
For the subsequent detail matting, models using only $\mathcal{L}_{DDC}$ or $\mathcal{L}_{t}$ both yield suboptimal performance. As shown in Fig.~\ref{fig:ab} (b), models using only $\mathcal{L}_{DDC}$ alone encounter artifacts in predictions, 
which aligns with our analysis in Sec.~\ref{sec:lt}. 
On the contrary, models using only $\mathcal{L}_{t}$ produce binary-like segmentation masks, lacking subtle details. 
Combining both terms yields optimal alpha mattes with sharp edges and natural textures, confirming their complementary roles. For more visualizations and ablations, please refer to the Supplementary.

\subsection{Scaling Study}

Here we investigate the scalability of our method on training data volumes. 
We augment the training set with images from a portrait matting dataset HHM50K~\cite{sun2023ultrahigh}, a portrait segmentation dataset Easyportrait~\cite{kvanchiani2023easyportrait}, and portrait photographs collected online. We train our model on varying amounts of data: 2K, 5K, 10K, 30K, 50K, and the full $90{,}533$ images. All images are resized proportionally to a minimum side length of $1080$, and each model is trained for $50$K iterations. 
All variants are evaluated on the P3M‑NP‑500 test set. As shown in Table~\ref{tab:scaling}, increasing data volume yields performance gains on both whole regions and boundaries, especially when moving from a low-data regime to large volumes (2K to 5K). Beyond 50K, performance tends to plateau, with additional data yielding only marginal improvements. 
Notably, the model trained with $90{,}533$ images achieves a lower SAD than the one trained solely on P3M-10K in our main experiment(Tab~\ref{tab:main result}, Row. 13). 
Considering the domain gap between the mixed training data (matting, segmentation, and web photographs) and P3M-10K, these results suggest that our method can benefit from  diverse training data of a reasonable scale, while still generalizing to P3M-10K.

\subsection{Complexity Analysis}
\begin{table}[!t]
\centering\noindent
\caption{\textbf{Scaling study on training data volume}. Our method shows consistent performance gains with data volume increase.}
\vspace{-18pt}
\begin{minipage}[t]{0.24\linewidth}\vspace{2pt}\centering
  \includegraphics[width=\linewidth]{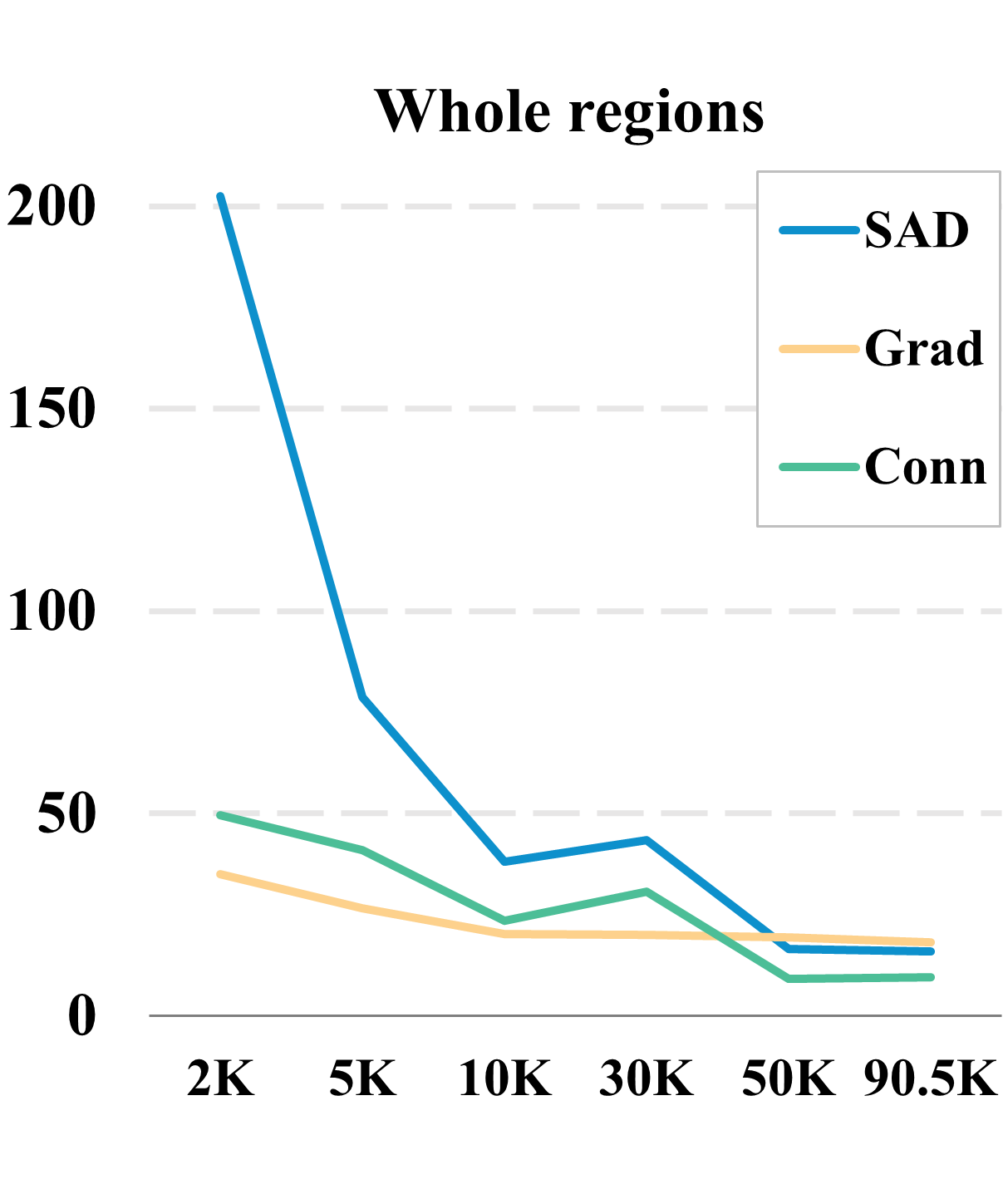}
\end{minipage}\hfill%
\begin{minipage}[t]{0.24\linewidth}\vspace{16pt}\centering
  \renewcommand{\arraystretch}{1.3}
  \resizebox{0.90\linewidth}{!}{%
    \begin{tabular}{lccc}
      \toprule
      Volume & SAD & Grad & Conn\\ \midrule
      2K & 202.53 & 35.11 & 49.60\\
      5K & 78.65  & 26.60 & 40.95\\
      10K& 38.00  & 20.26 & 23.44\\
      30K& 43.50  & 19.96 & 30.67\\
      50K& 17.14 & 19.33 & 10.46\\
      90.5K&16.60 & 19.38 & 9.24\\
      \bottomrule
    \end{tabular}}
\end{minipage}\hfill%
\begin{minipage}[t]{0.24\linewidth}\vspace{2pt}\centering
  \includegraphics[width=\linewidth]{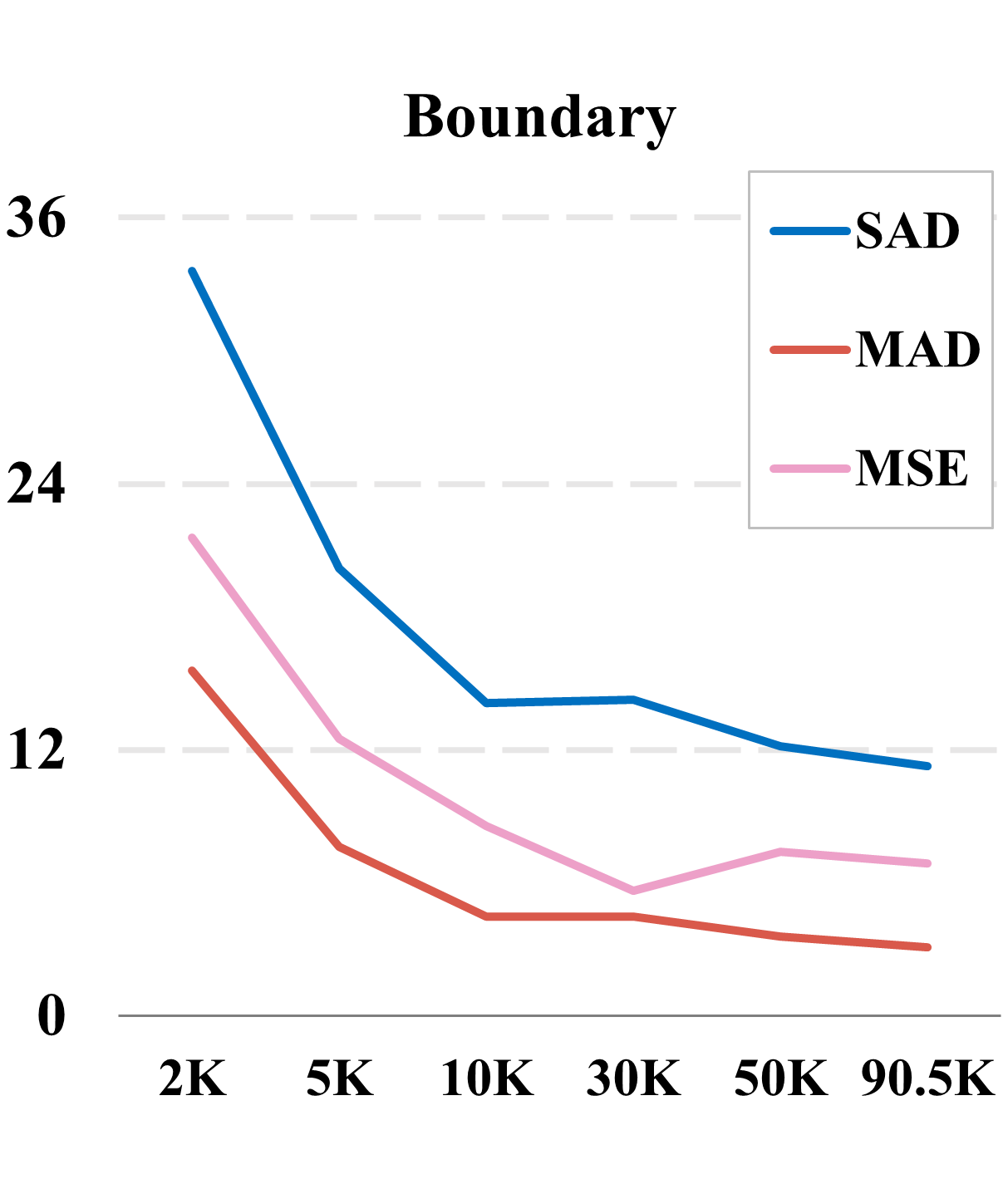}
\end{minipage}\hfill%
\begin{minipage}[t]{0.24\linewidth}\vspace{16pt}\centering
  \renewcommand{\arraystretch}{1.24}
  \resizebox{0.90\linewidth}{!}{%
    \begin{tabular}{lccc}
      \toprule
      Volume & SAD & MAD & MSE\\ \midrule
      2K & 33.59 & 15.56 & 21.58\\
      5K & 20.17 & 7.64  & 12.50\\
      10K& 14.11 & 4.48  & 8.55\\
      30K& 14.26 & 4.49  & 8.65\\
      50K& 12.17 & 3.60  & 7.41\\
      90.5K&11.74& 3.36  & 7.15\\
      \bottomrule
    \end{tabular}}
\end{minipage}
\label{tab:scaling}
\end{table}

\begin{table}[!t]
\centering\noindent
\caption{\textbf{Complexity analysis on params and training efficiency}. Our method has the smallest trainable parameters and is comutational efficient.}
\vspace{-12pt}
\begin{minipage}[t]{0.3\linewidth}\vspace{0pt}\centering
  \includegraphics[width=\linewidth]{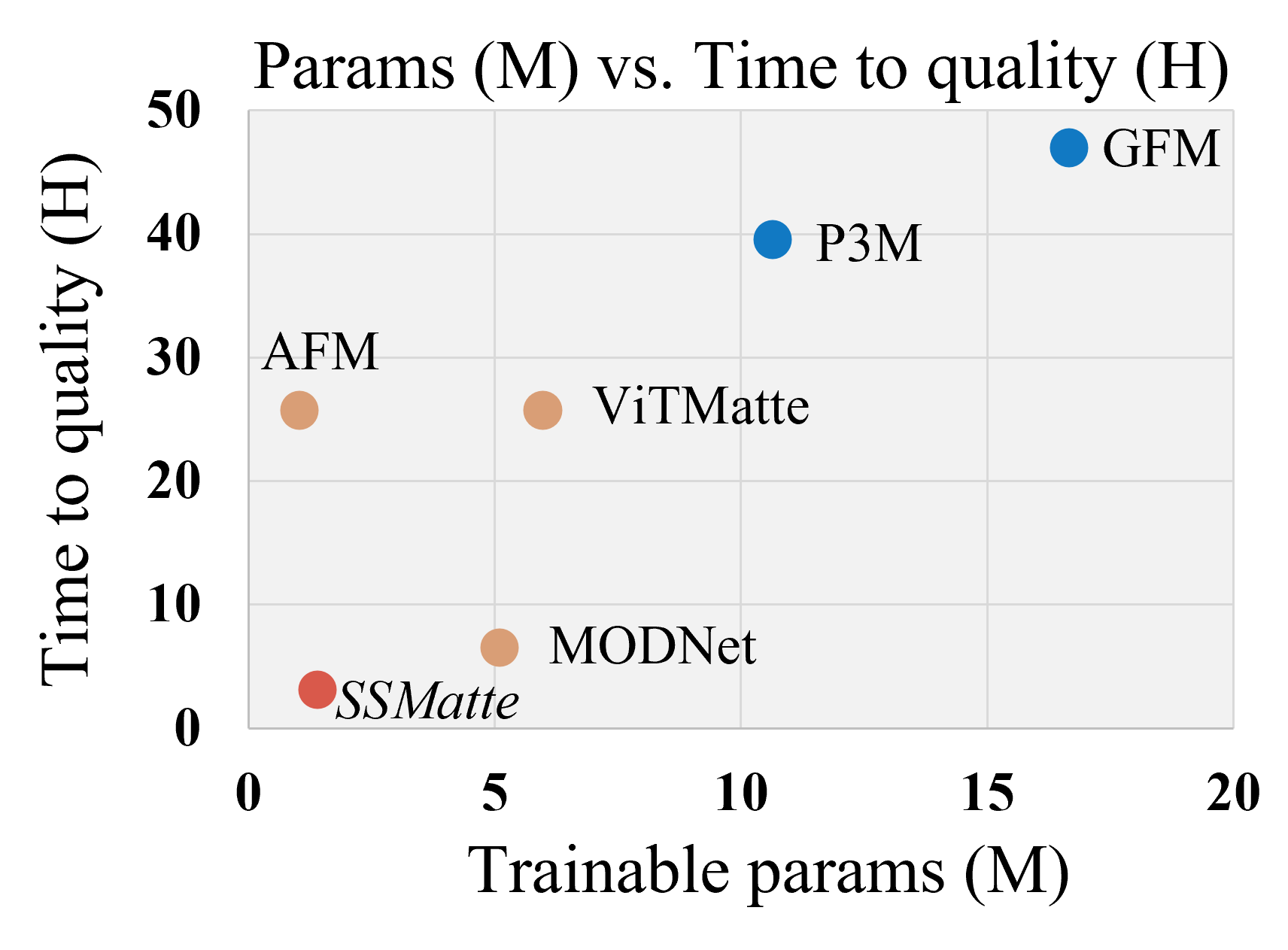}
\end{minipage}\hfill%
\begin{minipage}[t]{0.65\linewidth}\vspace{9pt}\centering
  \vspace{-8pt}
  \setlength{\tabcolsep}{3pt}
  \renewcommand{\arraystretch}{1.1}
  \resizebox{0.95\linewidth}{!}{%
    \begin{tabular}{lccccc}
      \toprule
      \multirow{2}{*}{Method} & \multirow{2}{*}{Label} & \multirow{2}{*}{GFLOPs} & \multirow{2}{*}{Params(M)} & Params & Time to quality\\
       & & & & (Trainable, M) & (3090 hours)\\
      \midrule
        MODNet~\cite{ke2022modnet}     & Matte & 187.73 & 6.48 & 6.48 & 5.10\\
        GFM~\cite{li2022bridging}        & Matte & 2210.25 & 46.96 & 46.96 & 16.67\\
        P3M~\cite{li2021privacy}        & Matte & 979.56 & 39.47 & 39.47 & 10.65\\
        ViTMatte-S~\cite{yao2024vitmatte} & Matte & 396.56 & 25.71 & 25.71 & 5.98\\
        AFM~\cite{liu2025training}        & Trimap & 396.56 & 25.71 & 25.71 & 1.04\\
        SSMatte (Ours)              & RGB & 534.07 & 25.16 & 3.10 & 1.41\\
      \bottomrule
    \end{tabular}}
\end{minipage}\hfill%
\label{tab:complex}
\end{table}

Here we evaluate the training efficiency of our method in terms of GFLOPs, number of parameters, and convergence speed. 
To quantify convergence speed, we report time-to-quality (TTQ), defined as the wall-clock time on one single RTX 3090 to first reach 
performance (SAD and SAD-T) within $10\%$ of the best achieved values on P3M-500-NP~\cite{li2021privacy} during the same training run and to remain below for three consecutive evaluations. 
Except for TTQ, all other metrics are measured with an input resolution of $1024 \times 1024$. As shown in Table~\ref{tab:complex}, our model has the smallest trainable parameters while achieving a fast time-to-quality. 
We also observe that non-fully supervised methods exhibit a significantly higher convergence speed, which mainly comes from the supervision signal. A near-controlled comparison is AFM vs. ViTMatte-S, which have similar GFLOPs, parameters, and model architectures, yet present a huge gap on TTQ.

\begin{table*}[!t]
\caption{\textbf{Automatic matting results on AIM-500~\cite{li2021deep}}. Backbones of AFM and SSMatte are initialized with DINOv2. $^\ddagger$ denotes pretraining on DUTS~\cite{wang2017learning}. Comp. denotes the mixed datasets of Distinction-646~\cite{qiao2020attention}, AM-2K~\cite{li2022bridging}, and Composition-1K~\cite{xu2017deep}. Transp. and Furni. are short for Transparent and Furniture, respectively. Best performances of fully supervised and non-fully supervised methods are in \textbf{boldface} and \underline{underline}, respectively. For full metrics please refer to the Supplementary. }
\centering
\vspace{-10pt}
\small
\setlength{\tabcolsep}{1pt}        

\resizebox{\textwidth}{!}{%
\begin{tabular}{lccccccccccccccccc}
\toprule
\multirow{2}{*}{Method} &
\multirow{2}{*}{Label} &
\multicolumn{3}{c}{Training set} &
\multicolumn{2}{c}{Average} &
\multicolumn{3}{c}{SAD-Type} &
\multicolumn{7}{c}{SAD-Category} \\
\cmidrule(lr){3-5}\cmidrule(lr){6-7}\cmidrule(lr){8-10}\cmidrule(lr){11-17}
& & Comp. & AM-2K & P3M-10K & SAD & Grad  & SO & STM 
& NS & Animal & Potrait & Transp. & Plant & Furni. & Toy & Fruit \\
\midrule
SHM$^\ddagger$~\cite{chen2018semantic}  & Matte & \ding{51} & & & 170.44 & 115.29  & 154.56 & 204.67 & 329.90 & 174.65 & 141.49 & 333.24 & 157.24 & 166.81 & 126.04 & 97.31\\
LF$^\ddagger$~\cite{zhang2019late}   & Matte & \ding{51} & & & 191.74 & 63.51 & 177.98 & 220.22 & 331.34 & 167.90 & 131.96 & 276.13 & 228.84 & 249.70 & 224.50 & 287.40\\
HATT$^\ddagger$~\cite{qiao2020attention} & Matte & \ding{51} & & & 479.17 & 238.63 & 509.75 & 338.11 & 270.07 & 579.96 & 484.85 & 264.35 & 433.96 & 299.19 & 447.01 & 104.73\\
GFM$^\ddagger$~\cite{li2022bridging} & Matte & \ding{51} & & & 52.66 & 46.11 & 35.45 & 123.15 & 181.90 & 28.18 & 27.61 & 190.50 & 75.77 & 80.94 & \underline{51.42} & \underline{27.87}\\
AIM$^\ddagger$~\cite{li2021deep}  & Matte & \ding{51} & & & \underline{43.92} & \underline{33.05} & \underline{31.80} & \underline{94.02} & \underline{134.31} & \underline{26.39} & \underline{24.68} & \underline{148.68} & \textbf{54.03} & \underline{62.70} & 53.15 & 37.17\\
SMat$^\ddagger$~\cite{ye2024unifying} & Matte & \ding{51} & & & \textbf{32.85} & \textbf{31.52} & \textbf{23.71} & \textbf{73.57} & \textbf{97.11} & \textbf{14.82} & \textbf{16.60} & \textbf{112.16} & \underline{63.39} & \textbf{41.75} & \textbf{31.33} & \textbf{22.25}\\
\midrule
AFM~\cite{liu2025training}  & Trimap & &\ding{51} & & 72.20 & 38.17 & 60.72 & 120.34 & \textbf{156.80} & 29.44 & 87.57 & \textbf{165.86} & \underline{69.51} & 174.99 & 67.12 & \underline{30.77}\\
AFM~\cite{liu2025training}  & Trimap & & &  \ding{51} & 67.03 & 40.03 & 50.44 & 128.09 & 200.58 & 46.98 & 29.41 & 234.01 & \textbf{67.06} & 113.96 & 62.23 & 82.15\\
SSMatte    & RGB    &  & \ding{51} & & 59.10 & 37.66 & 40.64 & 129.47 & 204.65 & \textbf{23.71} & 53.21 & 182.36 & 98.17 & 83.45 & 44.13 & 58.17\\
SSMatte     & RGB    & & & \ding{51} & 59.51 & 37.40 & 40.39 & 133.73 & 208.48 & 32.13 & \textbf{14.63} & 212.17 & 81.00 & 144.85 & 39.39 & 64.37\\
SSMatte$^\ddagger$    & RGB   &  & \ding{51} & & 52.41 & \underline{35.23} & 36.17 & \underline{113.55} & \underline{182.02} & 28.05 & 26.54 & \underline{175.47} & 73.85 & 97.83 & 42.35 & 50.34\\
SSMatte$^\ddagger$     & RGB  & &  & \ding{51} & \underline{48.74} & 35.32 & \underline{30.47} & \textbf{113.12} & 199.46 & 28.94 & 16.51 & 188.86 & 75.87 & \underline{67.58} &\underline{38.04} & 40.45\\
SSMatte$^\ddagger$    & RGB   &  & \ding{51} & \ding{51} & \textbf{45.78} & \textbf{33.98}  & \textbf{27.18} & 118.77 & 189.57 & \underline{25.68} & \underline{15.83} & 182.01 & 78.73 & \textbf{55.38} & \textbf{35.72} & \textbf{29.70}\\
\bottomrule
\label{tab:aim}
\end{tabular}%
}

\label{tab:AIM-500}
\end{table*}

\subsection{Generalization Study}
Here we evaluate the generalizability of our method across different foreground types (Salient Opaque, SO; Transparent/Meticulous, STM; Non-Salient, NS) and matting categories. 
Note that all baselines are pretrained on DUTS~\cite{wang2017learning} and trained on a mixed dataset, consisting of Distinction-646~\cite{qiao2020attention}, AM-2K~\cite{li2022bridging}, and Composition-1K~\cite{xu2017deep}. For a comprehensive comparison, we train our models on (1) AM-2K, (2)P3M-10K, and (3) a combination of AM-2K and P3M-10K, along with pretraining on DUTS. 
Different from AIM~\cite{li2021deep}, which requires fully supervised pretraining, we only need to pretrain the Semantic Inducer on DUTS. 
This pretraining is label-free, lightweight, and computationally efficient, as the Semantic Inducer comprises only $0.55$M parameters out of the total $3.10$M trainable parameters. 
As shown in Table~\ref{tab:aim}, with this simple pretraining, our method performs comparably with the fully supervised AIM~\cite{li2021deep}, and even surpasses it on certain categories. 
Notably, our method achieves state-of-the-art performance on portrait matting, even compared with fully supervised ones.  
Variants trained solely on the class-specific dataset show overfitting to the corresponding category, whereas pretraining on DUTS significantly mitigates this issue, improving performance across all object categories. 
This demonstrates that our method can effectively leverage diverse datasets to generalize to a wide range of matting targets. 
Among all the foreground types, our method performs best on SO foregrounds, achieving comparable performance with SMat~\cite{ye2024unifying}, while for STM and NS foregrounds, our method produces suboptimal results, comparable with GFM~\cite{li2022bridging}. This limitation aligns with our methodology, as \texttt{[CLS]} token naturally attends to salient objects. For more details, please refer to the Supplementary. 


\section{Conclusion}

We present a novel self-supervised training paradigm to liberate deep image matting from any manual annotations. 
In the proposed method, SSMatte, we mine a semantic matting prompt from self-supervised ViT features via a novel, training-efficient semantic anchoring loss, and utilize it to anchor a detail matting decoder based on the alpha-RGB consistency. 
Experiments demonstrate that our method outperforms prior weakly supervised methods and produces competitive results with previous SOTA fully supervised methods. 
Further study highlights the data efficiency and scalability of our method. 
For future work, we plan to extend our method to more challenging foreground types like STM and NS. 

\section{Acknowledgment}
This work is jointly supported by the National Natural Science Foundation of China under Grant No. 62576146 and the Hubei Provincial Natural
Science Foundation of China under Grant No. 2024AFB566.

\clearpage  

%
%
\bibliographystyle{splncs04}
\bibliography{main}
\end{document}